\documentclass{fairmeta} 

\usepackage{latexsym}
\usepackage{enumitem}
\usepackage{algorithm}
\usepackage{algorithmic}
\usepackage{amssymb}
\usepackage{amsmath}
\usepackage{array}
\usepackage{wrapfig}
\usepackage{colortbl} 
\usepackage{xspace}

\newcommand{\openskill}{\textsc{OpenSkill}\xspace}

\definecolor{tendergreen}{RGB}{235, 247, 235} 
\definecolor{tenderblue}{RGB}{237, 243, 252}  
\definecolor{openskillcol}{RGB}{214, 230, 252} 

\newcolumntype{R}{>{\centering\arraybackslash}p{1.41cm}}
\newcolumntype{K}{>{\columncolor{openskillcol}\centering\arraybackslash}p{1.41cm}}
\newcommand{\refcell}[1]{\textcolor{gray!75}{#1}}

\newcommand{\yes}{\checkmark}
\newcommand{\no}{\textcolor{gray!60}{$\times$}}

\definecolor{promptframe}{RGB}{52, 103, 170}   
\definecolor{promptback}{RGB}{235, 242, 255}   
\newtcolorbox{promptbox}[1]{%
  enhanced,
  colback=promptback, colframe=promptframe,
  colbacktitle=promptframe, coltitle=white, fonttitle=\bfseries,
  halign=left,
  title={#1}, boxrule=1pt, arc=2.5mm,
  left=2.5mm, right=2.5mm, top=2mm, bottom=2mm,
  toptitle=1mm, bottomtitle=1mm
}

\definecolor{takeawayframe}{gray}{0.20} 
\definecolor{takeawayback}{gray}{0.95}  
\newtcolorbox{takeaway}[1][]{%
  enhanced, unbreakable, before skip=6pt, after skip=6pt,
  colback=takeawayback, colframe=takeawayframe,
  colbacktitle=takeawayframe, coltitle=white, fonttitle=\bfseries\normalsize,
  title={Takeaway\ifstrempty{#1}{}{~#1}},
  boxrule=1pt, arc=1.5mm,
  left=2.5mm, right=2.5mm, top=1.5mm, bottom=1.5mm,
  toptitle=1mm, bottomtitle=1mm,
  fontupper=\normalsize
}

\newcommand{\afflogo}[2]{%
  \IfFileExists{logos/#1.pdf}{\raisebox{-0.5\height}{\includegraphics[height=0.5cm]{logos/#1.pdf}}}{%
   \IfFileExists{logos/#1.png}{\raisebox{-0.5\height}{\includegraphics[height=0.5cm]{logos/#1.png}}}{%
    \fcolorbox{gray!45}{white}{\rule[-2pt]{0pt}{8pt}\,\scriptsize #2\,}}}%
}

\title{\openskill: Open-World Self-Evolution for LLM Agents}

\author[1,*]{Zhiling Yan}
\author[1,*]{Dingjie Song}
\author[2]{Hanrong Zhang}
\author[1]{Wei Liang}
\author[3,4]{Yuxuan Zhang}
\author[5]{Yutong Dai}
\author[1]{Lifang He}
\author[2]{Philip S. Yu}
\author[5]{Ran Xu}
\author[6]{Xiang Li}
\author[1,\dagger]{Lichao Sun}

\affiliation[1]{Lehigh University}
\affiliation[2]{University of Illinois Chicago}
\affiliation[3]{University of British Columbia}
\affiliation[4]{Vector Institute}
\affiliation[5]{Salesforce AI Research}
\affiliation[6]{Massachusetts General Hospital and Harvard Medical School}

\contribution[*]{Equal contribution}
\contribution[\dagger]{Corresponding author}

\abstract{%
Self-evolving agents requires adaptation after deployment, but existing approaches assume a usable learning loop, such as curated skills, successful trajectories, or verifier signals. Real open-world deployments may provide none of these, offering only a task prompt.
In this work, we study \emph{open-world self-evolution}, where an agent must build both its skills and its own verification signals from scratch, using open-world resources but no target-task supervision.
We propose \openskill, a framework that bootstraps this loop: it acquires grounded knowledge and verification anchors from documentation, repositories, and the web, synthesizes them into transferable skills, and refines those skills against self-built virtual tasks grounded in the anchors rather than in target answers.
The open world thus supplies both the knowledge to be learned and a supervision-independent practice environment, with target-task supervision reserved for final evaluation.
Across three benchmarks and two target agents, \openskill attains the best automated pass rate while satisfying the no-supervision constraint.
Analysis shows its skills transfer across models without model-specific adaptation, and its self-built verifier aligns with ground-truth outcomes despite never accessing them.%
}

\metadata[Code]{\url{https://github.com/OpenLAIR/OpenSkill}}
\metadata[Website]{\url{https://openlair.github.io/openskill/}}

\hypersetup{
  pdftitle={OpenSkill: Open-World Self-Evolution for LLM Agents},
  pdfauthor={Zhiling Yan, Dingjie Song, Hanrong Zhang, Wei Liang, Yuxuan Zhang, Yutong Dai, Lifang He, Philip S. Yu, Ran Xu, Xiang Li, Lichao Sun}
}

\newcommand{\onelogo}[1]{\raisebox{-0.5\height}{\includegraphics[height=0.55cm]{logos/#1}}}
\renewcommand{\titlelogos}{%
  \onelogo{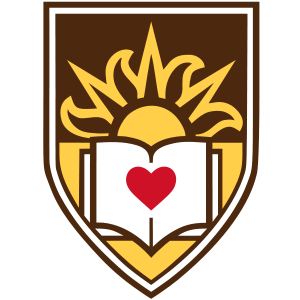}\;\onelogo{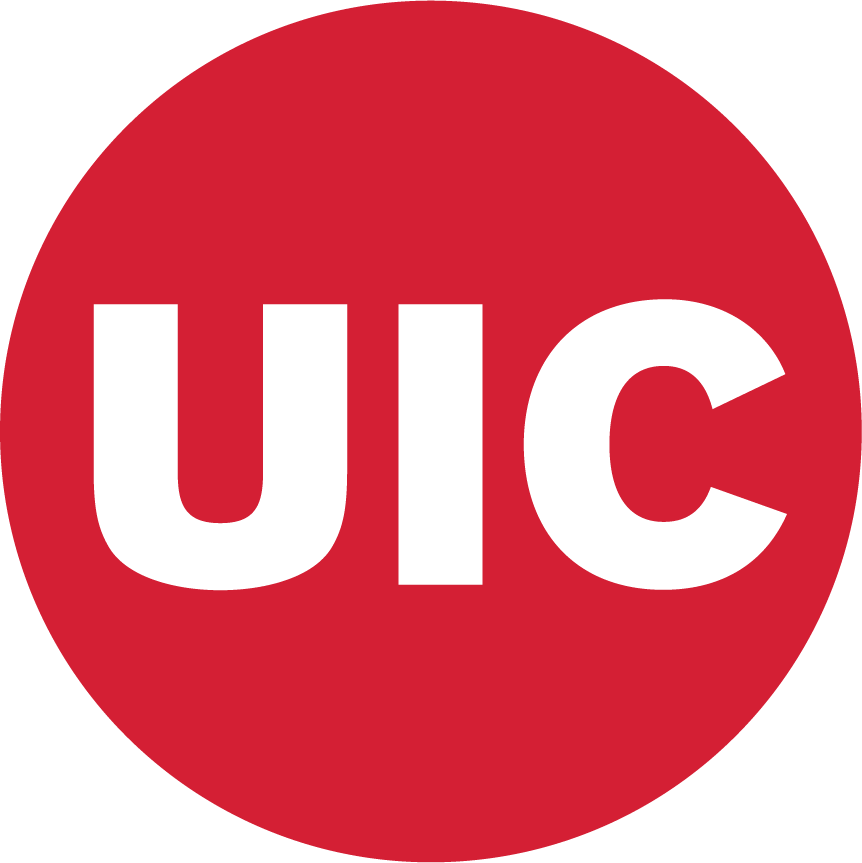}\;\onelogo{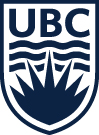}\;%
  \onelogo{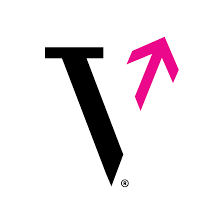}\;\onelogo{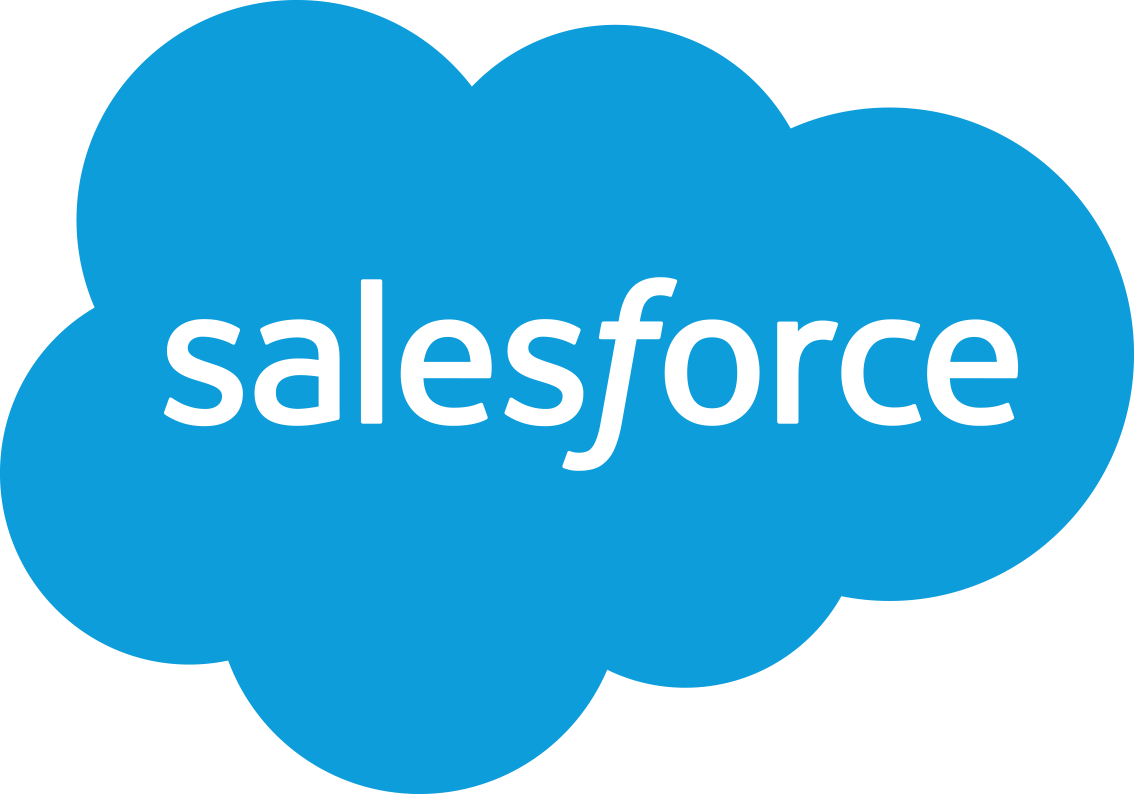}\;\onelogo{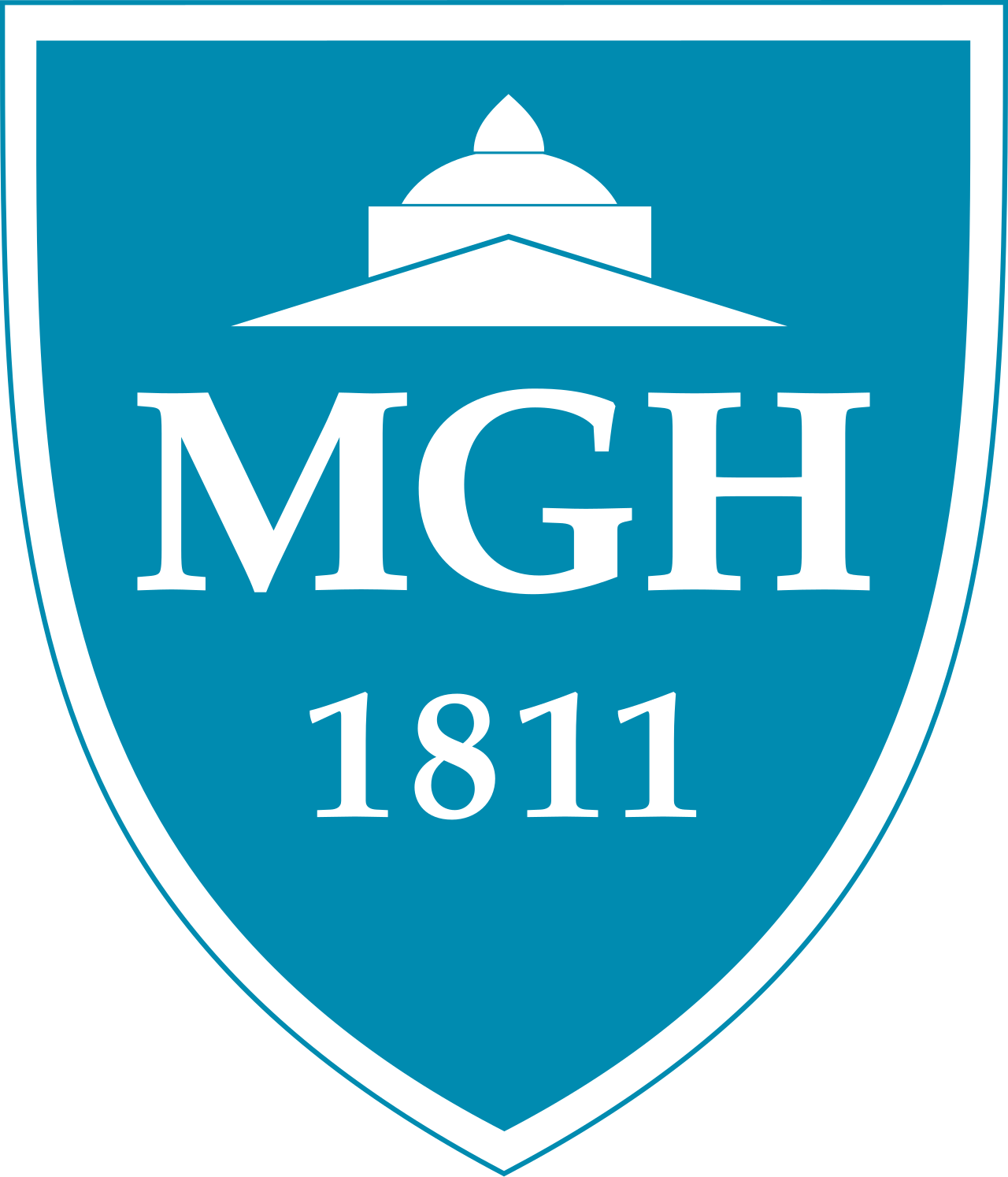}\;\onelogo{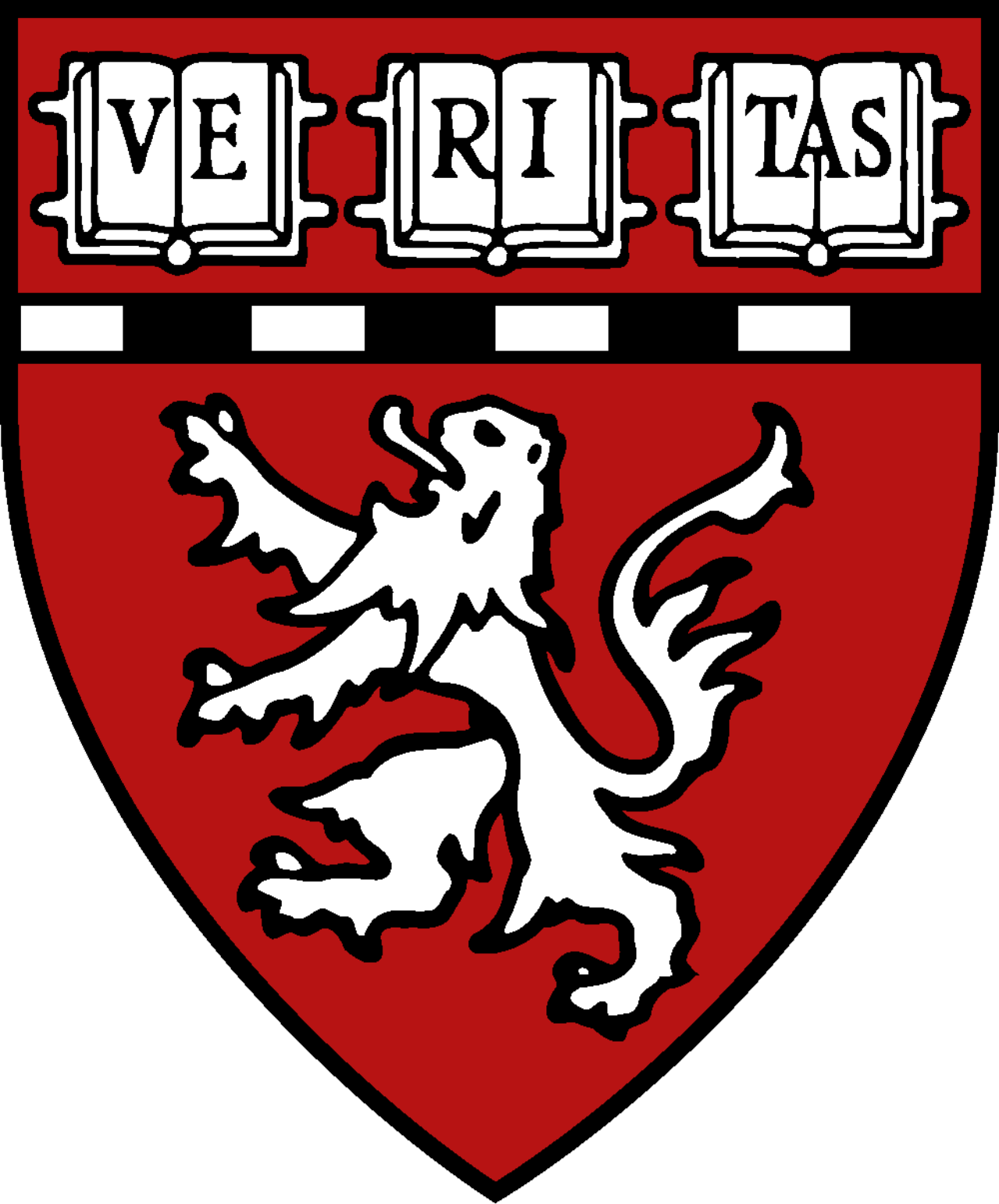}%
}

\begin{document}
\maketitle

\section{Introduction}

LLM agents can use tools and external resources to solve open-ended tasks beyond text generation \citep{react2022,toolformer2023,lats2023,webgpt2021,sweagent2024}.
Because such tasks change across environments, self-evolving agents aim to improve after deployment by accumulating reusable knowledge or behavior \citep{reflexion2023,voyager2023,agentfactory2026,skillrl2026}.

Existing self-evolving agents often assume a usable learning loop, such as curated skills, successful traces, or task feedback \citep{agentfactory2026,skillrl2026}.
Real open-world deployments may provide only a seed task prompt, with no initial skills or verifier for judging improvement.

\begin{figure}[t]
  \centering
  \includegraphics[width=\columnwidth]{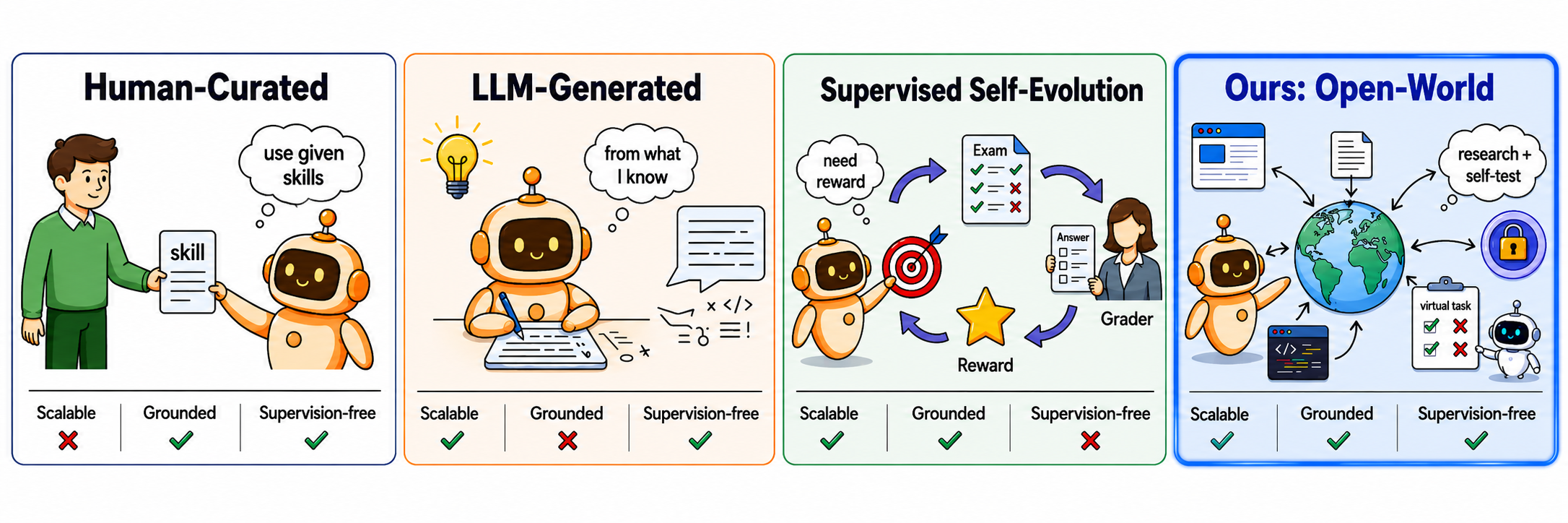}
  \caption{\textbf{Paradigms for self-evolving agent skills.}
  Unlike human-curated, LLM-generated, or supervised self-evolution paradigms,
  \openskill (ours) acquires skills from the open world and verifies them
  with self-built virtual tasks, making it simultaneously \emph{scalable},
  \emph{grounded}, and \emph{supervision-free}.}
  \label{fig:teaser}
\end{figure}

We call this capability \emph{open-world self-evolution} (Figure~\ref{fig:teaser}).
Here the agent starts from only a seed task prompt and open-world resources, and must build both its skills and its verification signals from scratch.
These resources include independently accessible evidence such as documentation, repositories, papers, tutorials, and web pages, but exclude hidden target answers, rewards, verifier outputs, or solution traces.
Facts alone are not enough: the same evidence must supply the learning loop itself.
This means building two coupled components: \emph{skill content} that captures what to learn, and a \emph{verification signal} that can improve those skills without target-task supervision.

\textbf{Limitation 1: skill construction.}
Existing approaches often rely on human-written skills, model-generated knowledge, or skills distilled from successful trajectories \citep{voyager2023,agentfactory2026,cascade2025,skillgen2026,autoskill}.
These sources are costly, bounded by prior knowledge, or unavailable before successful task attempts.
In an open-world setting, the agent must instead infer what to learn, acquire external evidence, and turn it into reusable skills.
This matters because many tasks require current or domain-specific knowledge, and recent benchmarks show that skill quality is often the limiting factor \citep{skillsbench2026,wildskills2026}.

\textbf{Limitation 2: verification construction.}
Existing self-improvement loops often revise behavior using task-level feedback, self-feedback, or verifier outputs \citep{reflexion2023,lats2023,selfrefine2023,agentfactory2026,coevoskills2026,sage2025}.
This works in curated benchmarks, but open-world deployment may expose no reliable feedback during learning.
The agent must therefore construct a separate practice environment whose supervision comes from open-world knowledge rather than hidden target-task answers.

This points to a sharper central question:
\begin{quote}
\emph{Can an LLM agent self-evolve in the open world?}
\end{quote}

To answer this, we propose \openskill, a framework for open-world self-evolution.
Given only a task prompt, a base model, tool access, and open-world resources, \openskill bootstraps a learning loop from scratch.
It proceeds in three stages: \emph{open-world knowledge acquisition} retrieves grounded knowledge and verification anchors from the open world; \emph{leakage-free skill evolution} drafts skills and refines them against self-built virtual tasks rather than target answers; and \emph{zero-shot target evaluation} deploys the refined skill to the target agent.
Open-world resources thus supply both the knowledge to be learned and a supervision-independent practice environment, and target-task supervision is reserved for final evaluation alone.
Empirically, \openskill delivers the best automated pass rate in every benchmark--agent setting (e.g., $+8.9$ / $+8.8$ over the strongest closed-world baseline on SkillsBench), transfers across models without adaptation, and builds a verifier covering 88.9\% of ground-truth test intents---all without target-task supervision during learning.
This paper makes three contributions:
\begin{itemize}[leftmargin=*]
    \item We define \emph{open-world self-evolution}: from only a task prompt, an agent must build both its skills and its own verification signals from open-world resources, with no target-task supervision.
    \item We propose \openskill, which bootstraps this loop and yields skills that transfer across models, with a practice environment for refining them.
    \item We show that \openskill achieves the best automated pass rate across three benchmarks and two model families, and transfers across models---without target-task supervision during learning.
\end{itemize}

\section{Open-World Self-Evolution}
\label{sec:method}

We introduce \emph{open-world self-evolution}, a setting in which an LLM agent
must improve from only a task prompt and open-world resources, with no initial
skills, demonstrations, rewards, or verifiers. We formalize it and its
supervision constraint (Section~\ref{sec:setup}), then present \openskill,
a three-stage pipeline that acquires open-world knowledge, refines skills against
self-generated virtual tasks, and deploys them zero-shot
(Section~\ref{sec:framework}).

\begin{figure}[t]
    \centering
    \includegraphics[width=1\linewidth]{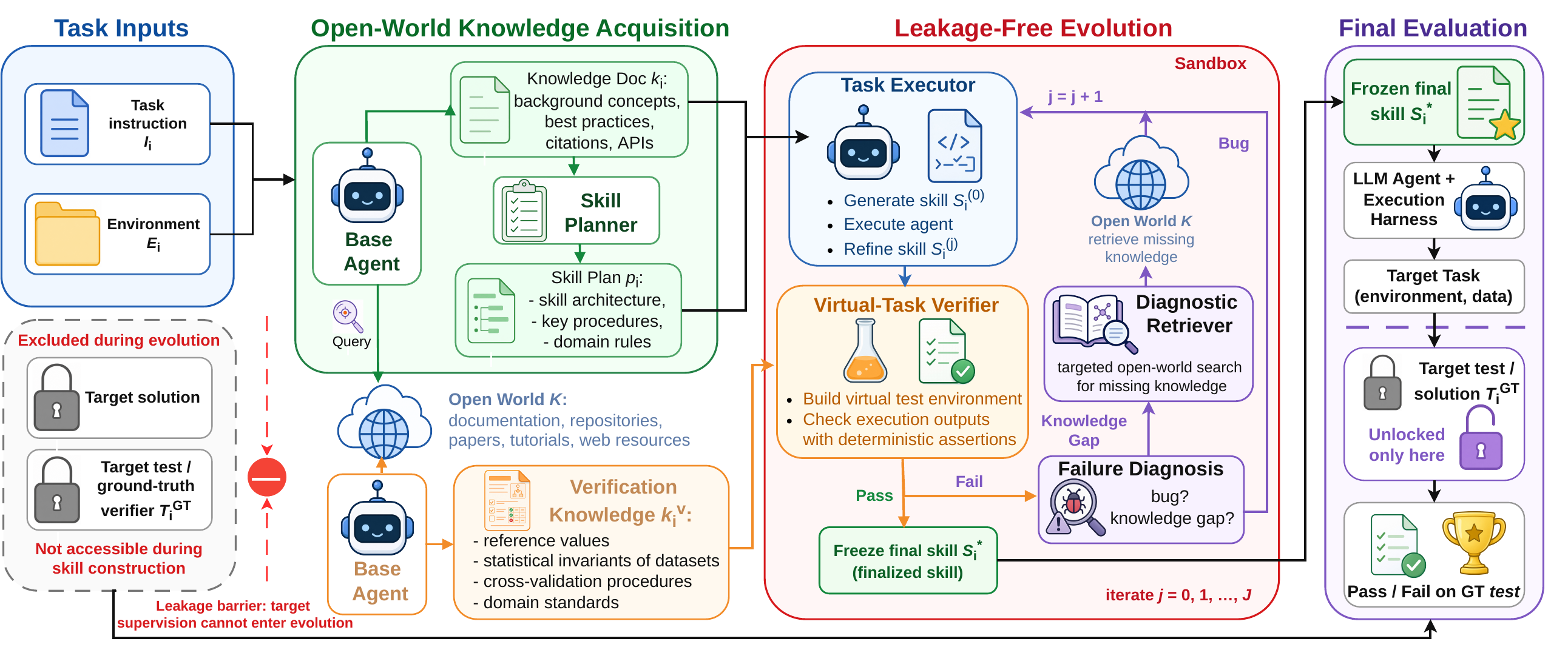}
    \caption{Overview of the \openskill framework. A base agent acquires open-world knowledge from external resources to build a skill plan, then iteratively generates, executes, and refines the skill in a sandbox, using a virtual-task verifier and diagnostic retriever to fix bugs and knowledge gaps. A leakage barrier keeps target supervision out of skill construction, unlocking it only for final evaluation.}
    \label{fig:architecture}
\end{figure}

\subsection{Problem Setting}
\label{sec:setup}

Consider a set of $n$ target tasks $\{(\mathcal{I}_i, \mathcal{E}_i)\}_{i=1}^{n}$, where $\mathcal{I}_i$ is a natural-language instruction and $\mathcal{E}_i$ is an execution environment. An LLM agent $\pi_\theta$ executes in $\mathcal{E}_i$ and produces a terminal state $x_i = \pi_\theta(\mathcal{I}_i, \mathcal{E}_i)$. Each task is paired with a ground-truth test suite $\mathcal{T}^{\text{GT}}_i \in \{0,1\}$.

An agent can self-evolve either by modifying its weights $\theta$ (e.g., via fine-tuning or reinforcement learning) or by augmenting its context with external knowledge artifacts. We adopt the latter, as it is computationally cheap, transferable across models, and inspectable, whereas weight modification is expensive, model-specific, and opaque. We formalize such an artifact as a \emph{skill} set $\mathcal{S}_i = \{s_{i,1}, \ldots, s_{i,m}\}$ that conditions the agent's behavior, $x_i = \pi_\theta(\mathcal{I}_i, \mathcal{S}_i, \mathcal{E}_i)$, without altering $\theta$. Self-evolution then reduces to constructing, for each task, a skill set under which the augmented agent passes its test suite: $\mathcal{T}^{\text{GT}}_i(\pi_\theta(\mathcal{I}_i, \mathcal{S}_i, \mathcal{E}_i)) = 1$.

The difficulty of the open-world setting is that this construction must proceed without target-task supervision. The agent observes only the instruction $\mathcal{I}_i$ and the environment $\mathcal{E}_i$; the ground-truth test suite $\mathcal{T}^{\text{GT}}_i$, reference solutions, and human feedback are hidden. Observing the input $(\mathcal{I}_i, \mathcal{E}_i)$ is not supervision: we reserve the term \emph{supervision} for dependence on the hidden signal $\mathcal{T}^{\text{GT}}_i$ (a gold answer, reward, or verifier output). Inspecting the input is thus permitted, while any dependence on $\mathcal{T}^{\text{GT}}_i$ is not. We call a construction procedure $f$ \emph{supervision-free} when it neither observes $\mathcal{T}^{\text{GT}}_i$ during skill construction nor reverse-engineers it to build the virtual tests used for refinement (Section~\ref{sec:stage2}). The skill set is then built purely from the observable input:
\begin{equation}
\hat{\mathcal{S}}_i = f(\mathcal{I}_i, \mathcal{E}_i).
\label{eq:construction-base}
\end{equation}

The observable input alone, however, is rarely sufficient to construct competent skills. We therefore let the agent interact with the open world and acquire open-world knowledge $\mathcal{K}$, such as public documentation, code repositories, papers, and tutorials, none of which reveals target-task supervision. The construction problem then expands to:
\begin{equation}
\hat{\mathcal{S}}_i = f(\mathcal{I}_i, \mathcal{E}_i, \mathcal{K}),
\label{eq:construction}
\end{equation}
where $f$ is the proposed \openskill pipeline.

\subsection{The \openskill Pipeline}
\label{sec:framework}

\openskill includes three stages
(Figure~\ref{fig:architecture}): acquiring domain knowledge from the open
world, refining skills against self-generated virtual tasks, and deploying the
final skill zero-shot to the target agent.
The stages are connected by the artifacts they pass on:
Stage~1 produces $k_i$, $p_i$, and $k_i^{v}$; Stage~2 drafts skills from
$(p_i, k_i)$ and refines them against virtual tests built from $k_i^{v}$,
emitting a frozen $\hat{\mathcal{S}}^*_i$ that Stage~3 deploys. A leakage
barrier keeps the hidden $\mathcal{T}^{\text{GT}}_i$ out of Stages~1--2.

\subsubsection{Stage 1: Open-World Knowledge Acquisition}
\label{sec:stage1}

Prior skill creation methods~\citep{skill-creator, autoskill}
construct skills entirely from the LLM's parametric knowledge
$\mathcal{K}_\theta$, the knowledge stored in the frozen weights $\theta$, as
opposed to the open-world knowledge $\mathcal{K}$ of
Section~\ref{sec:setup}. This limits the resulting skills to what the model
already knows, which is insufficient for tasks that require up-to-date APIs,
project-specific conventions, or niche domain rules.

\openskill expands the knowledge base by querying the open world. Given $(\mathcal{I}_i, \mathcal{E}_i)$, the pipeline first retrieves
task-relevant knowledge from the open world $k_i = \mathcal{D}(\mathcal{I}_i, \mathcal{K}),$ $ k_i \subset \mathcal{K},$
where $\mathcal{D}$ is an open-world retrieval function that traverses $\mathcal{K}$
and returns knowledge documents containing background concepts, best
practices, API documentation, and source citations. A structured skill plan $p_i$
is then synthesized based on $(\mathcal{I}_i, \mathcal{E}_i, k_i)$,
specifying the skill architecture, key procedures, and domain rules
(implementation in Appendix~\ref{app:retrieval}).

In addition to knowledge for skill construction, the pipeline retrieves
\emph{verification knowledge} $k_i^{v} = \mathcal{D}^{v}(\mathcal{I}_i, \mathcal{K}),$ $ k_i^{v} \subset \mathcal{K}$,
which provides independently verifiable anchors for later quality assessment,
including reference values from official documentation, statistical invariants
of well-known datasets, cross-validation procedures from domain standards, and
expected output formats. $k_i^{v}$ is used in Stage~2 to ground virtual test
generation (implementation in Appendix~\ref{app:surrogate}).

To prevent answer leakage, all queries issued by $\mathcal{D}$ and
$\mathcal{D}^{v}$ are filtered to exclude the benchmark name and any
identifiers that could lead to $\mathcal{T}^{\text{GT}}$; we audit this
information isolation in Appendix~\ref{app:leakage_audit}.

\subsubsection{Stage 2: Leakage-Free Skill Evolution}
\label{sec:stage2}

Given $(\mathcal{I}_i, \mathcal{E}_i, p_i, k_i)$, the base agent $\pi_\theta$ generates an initial
skill set $\hat{\mathcal{S}}^{(0)}_i = \{\hat{s}_{i,1}^{(0)}, \ldots,
\hat{s}_{i,m}^{(0)}\}$, whose size $m$ ($1\le m\le 4$) is fixed by the Stage-1
plan $p_i$, and refines all $m$ skills jointly within a single agent session.

To assess the skill without ground-truth feedback, the pipeline constructs a
virtual test suite $\tilde{\mathcal{T}}_i= \{\tilde{t}_{i,1}, \ldots, \tilde{t}_{i,K}\}$ grounded in the verification knowledge $k_i^v$ obtained in
Stage~1:
\begin{equation}
	\tilde{\mathcal{T}}_i = g(\mathcal{I}_i, \mathcal{E}_i, k_i^v)
,
\label{eq:virtual-gen}
\end{equation}
$\tilde{\mathcal{T}}_i$
serves as a proxy for $\mathcal{T}^{\text{GT}}_i$ to guide skill refinement.
The pipeline $f$ (Eq.~\ref{eq:construction}) calls $g$ internally: it scores
each round's skills with $g$'s tests to drive refinement, never observing
$\mathcal{T}^{\text{GT}}_i$.
Each virtual test $\tilde{t}_{i,k} \in \{0,1\}$ is a deterministic
assertion anchored to independently verifiable facts rather than guessing what the ground-truth tests might check. For example, it checks the known row count of a public dataset, the expected range of a standard metric, or the documented output format of a library function. The generator $g$ is realized as an isolated verifier LLM session that emits a deterministic \texttt{pytest} suite (Appendix~\ref{app:verifier}).

The pipeline iterates for up to $J$ rounds. At each round $j$, the current
skill set $\hat{\mathcal{S}}_i^{(j)}$ is executed and evaluated against $\tilde{\mathcal{T}}_i$. The virtual pass rate:
\begin{equation}
	\tilde{r}^{(j)} = \frac{1}{|\tilde{\mathcal{T}}_i|}
\sum_{k=1}^{K} \tilde{t}_{i,k}\!\bigl(\pi_\theta(\mathcal{I}_i,
\hat{\mathcal{S}}_i^{(j)}, \mathcal{E}_i)\bigr)
\label{eq:virtual-pass-rate}
\end{equation}
serves as a proxy for skill quality.
This proxy is reliable only if $\tilde{r}$ aligns with the hidden
$\mathcal{T}^{\text{GT}}_i$; otherwise refinement may reward skills that overfit
the virtual tests. We treat this alignment as an empirical question and measure
it directly in Section~\ref{sec:surrogate_analysis}.

	\textbf{Diagnostic-driven refinement.}
When $\tilde{r}^{(j)} < 1$,
the pipeline produces a structured failure diagnostic $\mathcal{F}^{(j)}$
comprising per-assertion results, root-cause analysis, and revision
suggestions, then refines the skill:
\begin{equation}
\hat{\mathcal{S}}^{(j+1)}_i = \pi_\theta(\hat{\mathcal{S}}^{(j)}_i,
\mathcal{F}^{(j)} \mid \mathcal{I}_i, \mathcal{E}_i, p_i, k_i).
\label{eq:refine}
\end{equation}

When the diagnostic indicates a \emph{knowledge gap} rather than an
implementation bug, the pipeline triggers a targeted retrieval
$k_i^{(\text{gap})} = \mathcal{D}(\mathcal{F}^{(j)}, \mathcal{K})$ and
injects the result into the refinement context. The gap-versus-bug decision is
made by an LLM classifier; we detail it and the targeted-retrieval budget in
Appendix~\ref{app:diagnosis}.

The loop terminates when $\tilde{r}^{(j)}=1$ or after at most $J$ refinement
rounds, whichever comes first; if $\tilde{r}^{(j)}<1$ throughout, it exhausts
the $J$-round budget (we set $J{=}3$). Auxiliary stall- and budget-based
early stops are detailed in Appendix~\ref{app:loop}.

\subsubsection{Stage 3: Zero-Shot Target Evaluation}
\label{sec:stage3}

After evolution, the final skill set $\hat{\mathcal{S}}^*_i$---the last refined
version at loop termination, edited in place rather than a best-of-$N$ snapshot
(Appendix~\ref{app:loop}, \ref{app:selection})---is deployed to the
target agent $\pi_{\theta'}$ in a zero-shot setting: the agent executes the target tasks
with $\hat{\mathcal{S}}^*_i$, and $\mathcal{T}_i^{\text{GT}}\in \{0,1\}$ determines pass or
fail. Because the skill is an explicit artifact rather than model weights, it
can be deployed to any target agent without retraining.

We evaluate the performance of $\pi_{\theta'}$ by the average pass rate on the hidden ground-truth
tests:
\begin{equation}
	\text{PassRate} = \frac{1}{n}\sum_{i=1}^{n}
\mathcal{T}^{\text{GT}}_i\!\bigl(\pi_{\theta'}(\mathcal{I}_i,
\hat{\mathcal{S}}^*_i, \mathcal{E}_i)\bigr).
\label{eq:eval}
\end{equation}
Note that the target agent $\pi_{\theta'}$ need not be the construction agent
$\pi_\theta$: because $\hat{\mathcal{S}}^*_i$ is a portable artifact, skills
built with one model can be deployed on another, and the hidden
$\mathcal{T}^{\text{GT}}_i$ enters the pipeline only here, at final evaluation.

\section{Experiment}

\subsection{Experimental Setup}
\label{sec:exp-setup}

\paragraph{Benchmarks.}
We evaluate on three agentic benchmarks. \textsc{SkillsBench}
\citep{skillsbench2026} is our primary benchmark, spanning 11 task domains where
skill quality is the limiting factor; \textsc{SocialMaze}
\citep{xu2025socialmaze} (social reasoning) and
\textsc{ScienceWorld}~\citep{wang2022scienceworld} (interactive science) add two
distinct task types. All are run under the open-world protocol of
Section~\ref{sec:method}, with the ground-truth tests $\mathcal{T}^{\text{GT}}_i$
hidden during construction and consulted only at final evaluation; full dataset
details are given in Appendix~\ref{app:datasets}.

\paragraph{Target agents.}
We report two target agents from different model families, Opus~4.6 (Claude
Code) and GPT~5.2 (Codex). For each, the pipeline is run end-to-end so that
skills are constructed and deployed with the same model.

\begin{table}[t]
\centering
\scriptsize
\setlength{\extrarowheight}{1.5pt}
\setlength{\tabcolsep}{3pt}

\begin{tabular}{l RRRRRR K @{\hspace{0.6em}} R}
\toprule
Domain
& \shortstack[c]{No Skill}
& \shortstack[c]{Self-Gen}
& \shortstack[c]{CoT}
& \shortstack[c]{Skill-Creator}
& \shortstack[c]{AutoSkill}
& \shortstack[c]{Memento}
& \shortstack[c]{\textbf{\openskill}}
& \shortstack[c]{\textcolor{gray!75}{Human}} \\
\midrule
\addlinespace[2pt]

\multicolumn{9}{l}{\textbf{Opus 4.6 (Claude Code)}} \\
\cmidrule(lr){1-9}
Software & 32.6 & 37.9 & 34.9 & \underline{51.3} & 36.0 & 34.4 & \textbf{59.9} & \refcell{38.8} \\
Office & 17.0 & 16.7 & 17.1 & 21.4 & 25.7 & \underline{31.4} & \textbf{50.0} & \refcell{50.0} \\
Science & 25.6 & 31.3 & 30.0 & \textbf{36.2} & 33.3 & \underline{35.0} & \underline{35.0} & \refcell{46.7} \\
Media & 36.1 & 27.9 & 20.4 & \underline{38.5} & 23.6 & 21.8 & \textbf{39.6} & \refcell{36.4} \\
Cybersecurity & 17.8 & 18.8 & 20.4 & 24.6 & 16.6 & \underline{28.8} & \textbf{44.1} & \refcell{55.0} \\
Finance & 17.5 & 16.7 & 20.0 & \textbf{27.5} & \underline{25.0} & \underline{25.0} & \underline{25.0} & \refcell{30.0} \\
Robotics & 27.6 & 13.3 & 16.0 & \textbf{36.0} & 4.0 & \underline{32.0} & \textbf{36.0} & \refcell{36.0} \\
Energy & \underline{41.2} & 11.1 & 40.0 & \textbf{60.0} & 33.3 & \textbf{60.0} & \textbf{60.0} & \refcell{66.7} \\
Manufacturing & 0.0 & 0.0 & 0.0 & 0.0 & 0.0 & 0.0 & 0.0 & \refcell{46.7} \\
Health & 24.8 & 19.8 & 19.2 & \underline{31.2} & 14.5 & 25.0 & \textbf{69.6} & \refcell{80.0} \\
Math & \underline{43.2} & 30.0 & 30.0 & \textbf{50.0} & 0.0 & 30.0 & \textbf{50.0} & \refcell{50.0} \\
\cmidrule(lr){1-9}
\rowcolor{tenderblue}
\textbf{Overall} & 25.5 & 23.9 & 23.9 & \underline{34.7} & 24.7 & 30.1 & \textbf{43.6} & \refcell{44.5} \\
\addlinespace[1pt]
\textit{$\Delta$ vs.\ No Skill} & -- & $-1.6$ & $-1.6$ & $+9.2$ & $-0.8$ & $+4.6$ & \textbf{$+18.1$} & \refcell{$+19.0$} \\
\midrule
\addlinespace[2pt]

\multicolumn{9}{l}{\textbf{GPT 5.2 (Codex)}} \\
\cmidrule(lr){1-9}
Software & 33.2 & \underline{48.4} & 47.2 & 44.4 & 16.7 & 19.5 & \textbf{49.1} & \refcell{42.5} \\
Office & \underline{32.9} & 31.0 & 26.2 & 26.2 & 9.4 & 14.3 & \textbf{44.3} & \refcell{48.6} \\
Science & \underline{30.4} & 30.3 & 29.8 & 21.9 & 5.5 & 13.8 & \textbf{48.6} & \refcell{48.3} \\
Media & \underline{31.3} & 31.0 & \textbf{31.8} & 30.9 & 15.2 & 18.2 & 30.4 & \refcell{58.2} \\
Cybersecurity & 25.0 & 20.8 & 34.7 & \underline{36.8} & 4.1 & 12.5 & \textbf{52.5} & \refcell{42.5} \\
Finance & 0.0 & \textbf{29.2} & \underline{25.0} & 20.8 & 8.4 & 12.5 & \underline{25.0} & \refcell{27.5} \\
Robotics & 16.0 & \underline{26.7} & \textbf{40.0} & 20.0 & 13.4 & 26.6 & \textbf{40.0} & \refcell{40.0} \\
Energy & 0.0 & 33.3 & \underline{55.6} & 22.2 & 11.0 & 22.3 & \textbf{80.0} & \refcell{53.3} \\
Manufacturing & 0.0 & \textbf{11.1} & 0.0 & 0.0 & 0.0 & 0.0 & 0.0 & \refcell{0.0} \\
Math & 30.0 & 33.3 & 33.3 & \textbf{50.0} & \underline{33.5} & 0.0 & \textbf{50.0} & \refcell{40.0} \\
Health & \underline{29.2} & \textbf{30.2} & \textbf{30.2} & 24.3 & 20.0 & 16.5 & 27.9 & \refcell{90.0} \\
\cmidrule(lr){1-9}
\rowcolor{tenderblue}
\textbf{Overall} & 25.0 & 32.2 & \underline{33.3} & 29.2 & 11.2 & 15.6 & \textbf{42.1} & \refcell{44.8} \\
\addlinespace[1pt]
\textit{$\Delta$ vs.\ No Skill} & -- & $+7.2$ & $+8.3$ & $+4.2$ & $-13.8$ & $-9.4$ & \textbf{$+17.1$} & \refcell{$+19.8$} \\
\bottomrule
\end{tabular}
\caption{Main results on SkillsBench (11 domains): average reward (pass rate, \%) by domain for two target agents. Best automated method per row in \textbf{bold}, second best \underline{underlined}; the \openskill column is shaded. \emph{Human} is a reference upper bound (set off on the right) and is excluded from the best-method comparison. The \textit{$\Delta$ vs.\ No Skill} row gives each method's overall pass-rate gain over the No-Skill floor (in points); negative values fall below it.}
\label{tab:main-results}
\end{table}

\paragraph{Baselines.}
We compare against seven automated baselines, all \emph{closed-world}---none
retrieves open-world knowledge or builds a self-verifier: \textbf{No Skill};
\textbf{Self-Gen} and \textbf{CoT}, which self-generate skills in a single pass
from parametric knowledge (CoT adds a structured chain-of-thought workflow);
\textbf{Skill Creator}~\citep{skill-creator} and
\textbf{AutoSkill}~\citep{autoskill}, skill-synthesis methods that iteratively
refine skills from prior knowledge or interaction traces; and
\textbf{Memento}~\citep{zhou2026memento}, a memory-/experience-based baseline
(plus \textbf{SkillNet}~\citep{liang2026skillnet} on \textsc{ScienceWorld}). A
\textbf{Human} upper bound is excluded from the comparison.
Appendix~\ref{app:baselines} details the baselines.

\paragraph{Metric and protocol.}
All methods share the same target agents and report average reward on the hidden test suite. A per-task cost breakdown of the \openskill pipeline (skill
creation vs.\ evaluation) is reported in Appendix~\ref{sec:cost}. Detailed baseline definitions, prompts, the
evaluation protocol, and full implementation configuration are deferred to
Appendices~\ref{app:baselines}, \ref{app:prompts}, \ref{app:eval-metrics},
\ref{app:hparams}, and~\ref{app:impl}.

\subsection{Main Results}

Table~\ref{tab:main-results} reports the average reward (pass rate) on
SkillsBench per domain and overall, for each target agent (exact model versions
in Appendix~\ref{app:models}).

\noindent\textbf{Best overall pass rate on both agents.}
\openskill achieves the best automated overall pass rate on both target
agents---43.6\% on Opus~4.6 and 42.1\% on GPT~5.2---beating the strongest
baseline by $+8.9$ and $+8.8$ points (Skill-Creator and CoT, respectively) and
landing within 1--3 points of the \emph{Human} upper bound (44.5\% / 44.8\%). It
is also the only method strong on both agents: the single-pass variants
(Self-Gen, CoT) help on GPT~5.2 but not Opus~4.6, while the iterative methods do
the reverse and collapse on GPT~5.2 (AutoSkill 24.7\%$\to$11.2\%, Memento
30.1\%$\to$15.6\%, both below the 25.0\% no-skill agent). Open-world acquisition
and leakage-free verification thus help regardless of the underlying model.

\noindent\textbf{Broad per-domain gains.}
\openskill is best or tied-best in 8 of 11 domains on Opus~4.6 and 7 of
11 on GPT~5.2, with the largest gains in knowledge-intensive domains (Opus:
Health 69.6\%, Software 59.9\%; GPT: Energy 80.0\%, Cybersecurity 52.5\%, both
above Human). It ties for second in Science and Finance on Opus~4.6, and
Manufacturing collapses to 0.0\% for all automated methods---a failure
open-world acquisition alone does not resolve.

\subsection{Beyond SkillsBench: Other Task Types}
\label{sec:additional-benchmarks}

We repeat the evaluation on SocialMaze and ScienceWorld
with the same two agents and closed-world baselines
. As Table~\ref{tab:additional-benchmarks} shows,
\openskill is the best automated method in all four columns---SocialMaze
82.7\% (Opus) / 70.7\% (GPT) and ScienceWorld 90.0\% / 85.3\%---improving over
the strongest baseline by $+0.9$ to $+2.2$ points, with larger gains on GPT~5.2
as on SkillsBench. A per-subtask SocialMaze breakdown is in
Appendix~\ref{app:socialmaze-detail}.

\begin{table}[tb]
\centering
\footnotesize
\setlength{\extrarowheight}{2.5pt}
\setlength{\tabcolsep}{4pt}
\begin{tabular}{l cc cc}
\toprule
& \multicolumn{2}{c}{\textbf{SocialMaze}} & \multicolumn{2}{c}{\textbf{ScienceWorld}} \\
\cmidrule(lr){2-3}\cmidrule(lr){4-5}
Method & Opus 4.6 & GPT 5.2 & Opus 4.6 & GPT 5.2 \\
\midrule
No Skill       & \underline{81.6} & 66.6 & 87.1 & 78.0 \\
Self-Gen       & 80.4 & 67.6 & 88.0 & 75.9 \\
CoT            & 79.4 & 66.7 & 88.0 & 80.0 \\
Skill Creator  & 81.0 & \underline{69.8} & 88.3 & \underline{83.1} \\
AutoSkill      & 77.2 & 65.9 & --   & --   \\
Memento        & 80.3 & 67.4 & --   & --   \\
SkillNet       & --   & --   & \underline{88.7} & 77.8 \\
\midrule
\rowcolor{tenderblue}
\openskill & \textbf{82.7} & \textbf{70.7} & \textbf{90.0} & \textbf{85.3} \\
\bottomrule
\end{tabular}
\caption{Average reward (\%) on SocialMaze and ScienceWorld for both target
agents. Best automated score per column in \textbf{bold}, second best
\underline{underlined}; ``--'' marks methods not run on a benchmark.}
\label{tab:additional-benchmarks}
\end{table}

\section{Analysis}
\label{sec:analysis}

We probe \openskill along three questions:\\
\textbf{RQ1 (Transferability)}: Do its skills transfer across models without
model-specific adaptation?\\
\textbf{RQ2 (Verifier quality)}: Without ground-truth tests, do the virtual
verifier's proxy tests align with and cover ground-truth test intents?\\
\textbf{RQ3 (Component contribution)}: How much does each design element
contribute?

\subsection{RQ1: Skill Generalization}
\label{sec:skill-generalization}

\begin{wrapfigure}{r}{0.5\textwidth}
  \centering
  \vspace{-\intextsep}
  \includegraphics[width=0.48\textwidth]{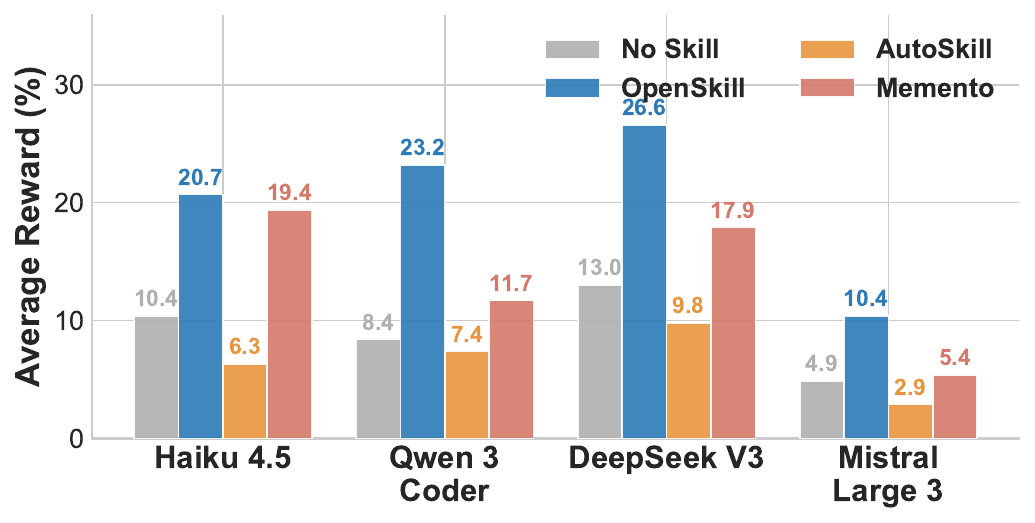}
  \caption{Average reward (\%) when transferring Opus 4.6-generated skills to other models on SkillsBench.}
  \label{fig:transfer}
\end{wrapfigure}

A key advantage of explicit, reusable skills is that skills generated by one model can be transferred to other models without retraining or regeneration. We evaluate this cross-model transferability by deploying the skill libraries produced by Opus 4.6 under three generation methods---\openskill (ours), AutoSkill, and Memento---directly onto four weaker models: Haiku 4.5,\footnote{\url{https://www.anthropic.com/news/claude-haiku-4-5}} Qwen 3 Coder,\footnote{\url{https://qwenlm.github.io/blog/qwen3-coder/}} DeepSeek V3~\citep{liu2024deepseek}, and Mistral Large 3.\footnote{\url{https://mistral.ai/news/mistral-3}} Each model is evaluated on SkillsBench.

Figure~\ref{fig:transfer} shows that \openskill-generated skills consistently yield the highest reward across all four target models, improving over the no-skill baseline by 5.5\%--14.8\%  points. Notably, these gains are achieved without any model-specific adaptation: the same skill files produced by Opus 4.6 are used as-is. Memento skills also transfer reasonably well to four models. AutoSkill performs worse than the no-skill baseline, suggesting that its generated skills are tightly coupled to the originating model and fail to generalize.

\begin{takeaway}[(RQ1)]
\openskill encodes task-relevant knowledge in a model-agnostic form, so the same
skill files transfer effectively across models---without any model-specific
adaptation---even to substantially weaker ones.
\end{takeaway}

\subsection{RQ2: Virtual Verifier Quality}
\label{sec:surrogate_analysis}

A central design element of \openskill is the \emph{Virtual Verifier}---a separate LLM agent that generates proxy test suites from task specifications alone, without access to ground-truth (GT) tests.
This proxy provides iterative feedback to the task executor during open-loop deployment where no GT oracle is available.
We evaluate the quality of surrogate-generated tests along two axes: \emph{alignment} with GT evaluation outcomes, and \emph{coverage} of GT test intents.

\begin{wraptable}{r}{0.5\textwidth}
\centering
\vspace{-\intextsep}
\footnotesize
\setlength{\extrarowheight}{2.5pt}
\begin{tabular}{l cc|c}
\hline
\hline
& \textbf{Reward > 0} & \textbf{Reward = 0} & \textbf{Total} \\
\hline
\textbf{Proxy Pass} & \cellcolor{tendergreen} 39.29\% & 29.76\% & 69.05\% \\
\textbf{Proxy Fail} & 9.52\%  & \cellcolor{tendergreen} 21.43\% & 30.95\% \\
\hline
\textbf{Total}   & 48.81\% & 51.19\% & 100.00\% \\
\hline
\hline
\end{tabular}
\caption{Percentage distribution of proxy results and GT rewards ($N=84$).}
\label{tab:sv-reward-proportions}
\end{wraptable}

\paragraph{Alignment with GT outcomes.}
Table~\ref{tab:sv-reward-proportions} reports the agreement between
virtual verifier decisions and GT evaluation outcomes.
The virtual verifier achieves 56.9\% precision and 80.5\% recall,
with an overall agreement rate of 60.7\%.
The association between virtual verifier decisions and GT reward
is statistically significant (Fisher's exact test $\mathrm{OR} = 2.97$, $p = 0.035$; point-biserial
$r = 0.242$, $p = 0.027$), confirming that the virtual verifier
provides a meaningful quality signal despite operating without
access to ground-truth tests. We analyze the remaining disagreement
cases and their failure modes in Appendix~\ref{app:failure-modes}.

\paragraph{Coverage of GT test intents.}
We further analyze how well the virtual verifier's generated tests
cover the evaluation intents of human-authored GT tests.
We randomly sample 15 tasks and use an LLM judge, i.e., opus 4.6 to
perform semantic matching between each GT test function
and the surrogate test suite, determining whether the
GT test's purpose (e.g., ``check output file exists,''
``verify numerical accuracy within tolerance'')
is addressed by at least one virtual test.

Across the 15 sampled tasks, the virtual verifier covers
88.9\% of GT test intents (120 out of 135).
The uncovered 11.1\% cluster in two categories:
(1)~anti-cheat meta-validation checks specific to the
benchmark infrastructure, and
(2)~deep semantic quality properties
(e.g., taxonomy coherence, lemmatization correctness)
that require domain expertise beyond what the task
specification provides.
Meanwhile, the virtual verifier generates a median of
3.4$\times$ more test functions per task than the GT suite, contributing 15.3 additional assertions
per task on average---primarily defensive checks on output
format, type validity, and domain-specific boundary conditions.

\begin{takeaway}[(RQ2)]
The virtual verifier uses no ground-truth tests, yet its proxy tests cover
most human-authored test intents and track the true outcomes closely
enough to gate skill generation on their own.
\end{takeaway}

\subsection{RQ3: Ablation Studies}
\label{sec:ablation}

\begin{wrapfigure}{r}{0.5\textwidth}
  \centering
  \vspace{-14mm}
  \includegraphics[width=0.48\textwidth]{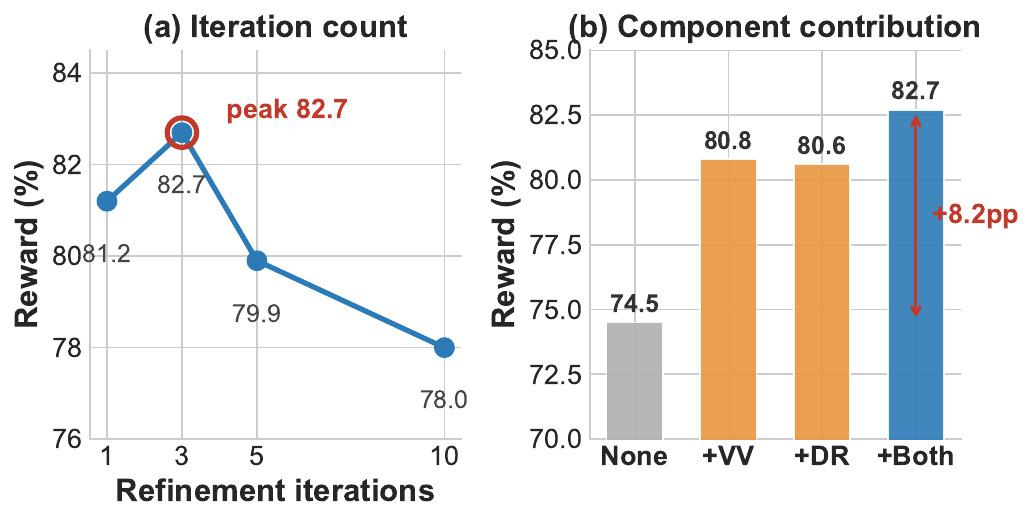}
  \caption{Ablations on \textsc{SocialMaze} (Opus 4.6). (a) Reward peaks at a few
  refinement iterations and degrades with more, indicating overfitting to virtual
  feedback. (b) Open-world query (DR) and the virtual verifier (VV) each improve
  over the parametric-only baseline and are largely complementary, with the
  combination performing best.}
  \label{fig:ablation}
  \vspace{-14mm}
  \end{wrapfigure}
We conduct two ablation studies on \textsc{SocialMaze} using Opus 4.6 to isolate the contributions of key design choices.

\paragraph{Iteration count.}
The virtual verifier loop refines each generated strategy through iterative critique--revision cycles. We vary the maximum number of iterations across $\{1, 3, 5, 10\}$ to examine its effect on downstream task performance. Figure~\ref{fig:ablation}a reports the results. Performance peaks at 3 iterations (82.7\%), matching the default configuration, and degrades slightly with additional iterations (79.9\% at 5, 78.0\% at 10), suggesting that excessive refinement introduces overfitting to virtual test feedback.

\paragraph{Component contribution.}
We ablate two core components of the pipeline: (1) the \emph{open world query} module, which retrieves external domain knowledge to inform strategy generation, and (2) the \emph{virtual verifier}, which provides proxy test feedback for iterative refinement. Figure~\ref{fig:ablation}b reports the results under four configurations. Removing both components yields 74.5\%, establishing a lower bound where the model relies solely on parametric knowledge for strategy generation. Adding either component individually recovers most of the performance: open world query alone contributes +6.1 percentage points (80.6\%), while the virtual verifier alone contributes +6.3 percentage points (80.8\%). Their contributions are largely complementary---combining both achieves 82.7\%, a further +2.1 percentage points over the better individual component---though the marginal gain of each is smaller in the presence of the other, indicating partial overlap in the errors they correct.

\begin{takeaway}[(RQ3)]
Open-world query and the virtual verifier each contribute substantially over a
parametric-only baseline and are largely complementary; refinement helps only
up to a point, after which additional iterations overfit to virtual feedback.
\end{takeaway}

\section{Related Work}

\subsection{Self-Evolving Agents and Agent Skills}

\begin{wraptable}{r}{0.5\textwidth}
\centering
\vspace{-\intextsep}
\footnotesize
\setlength{\extrarowheight}{2pt}
\setlength{\tabcolsep}{3pt}
\begin{tabular}{l cccc}
\toprule
Method & OW retr. & Refine & SF verif. & Artifact \\
\midrule
No Skill      & \no  & \no  & \no  & \no  \\
Self-Gen      & \no  & \no  & \no  & \yes \\
CoT           & \no  & \no  & \no  & \yes \\
Skill Creator & \no  & \yes & \no  & \yes \\
AutoSkill     & \no  & \yes & \no  & \yes \\
Memento       & \no  & \yes & \no  & \yes \\
\rowcolor{tenderblue}
\openskill (ours) & \yes & \yes & \yes & \yes \\
\bottomrule
\end{tabular}
\caption{Capability comparison of the automated methods. \emph{OW retr.}:
acquires open-world knowledge beyond parametric/experience memory;
\emph{Refine}: iteratively refines skills; \emph{SF verif.}: builds a
supervision-free verification signal (no target-task feedback); \emph{Artifact}:
produces an explicit, model-transferable skill. \textsc{SkillNet}
(\textsc{ScienceWorld} only) is omitted.}
\label{tab:capability}
\end{wraptable}

LLM agents interleave reasoning and actions \citep{react2022}, teach themselves to call tools \citep{toolformer2023}, and improve planning through structured deliberation \citep{treeofthoughts2023,lats2023}.
To improve after deployment, self-evolving agents accumulate reusable knowledge through reflection over past attempts \citep{reflexion2023}, executable skills learned by exploration \citep{voyager2023}, subagents distilled from successful solutions \citep{agentfactory2026}, and cumulative skill creation \citep{cascade2025}; recent work couples skill learning with verification, co-evolving skills and their verifiers \citep{coevoskills2026}.
Reinforcement-learning variants instead internalize skills into model weights \citep{skillrl2026,sage2025,skillzero2026}, yielding knowledge that is hard to inspect, edit, or transfer across models.
A growing body treats skills as a managed resource: benchmarks show self-generated skills are unreliable \citep{skillsbench2026,agentskills2026}, while retrieval, compression, and multi-objective selection improve deployment \citep{skillflow2025,graphskills2026,skillreducer2026,effiskill2026,skillmoo2026}; skills can further be viewed as structured prompts, linking to automatic prompt optimization \citep{ape2022,opro2023}.
Unlike these lines, \openskill makes open-world acquisition the primary source of skill content, keeps skills as explicit and transferable artifacts rather than model-bound behaviors, and refines them without target-task supervision (Table~\ref{tab:capability}); better retrieval and compression remain complementary at deployment.

\subsection{Open-World Knowledge Acquisition}
Retrieval-augmented generation grounds outputs in external, non-parametric evidence \citep{rag2020,ragsurvey2024}, and browser-assisted, deep-research, and environment-interactive agents search the web, repositories, and tools to complete knowledge-intensive, long-horizon tasks \citep{webgpt2021,deepresearch2025,webarena2023,sweagent2024}.
These methods retrieve knowledge to answer a query or complete a single task, whereas \openskill uses open-world retrieval as the substrate for synthesizing persistent, reusable skills and for grounding a self-verification signal.

\subsection{Self-Verification and Self-Generated Evaluation}
Without target-task supervision, an agent must judge its own outputs: prior work aggregates multiple reasoning paths \citep{selfconsistency2022}, iterates on self-feedback \citep{selfrefine2023}, and uses LLMs as judges \citep{mtbench2023,llmjudge2024}, while in code domains self-generated tests filter or repair solutions through execution feedback \citep{codet2022,selfdebug2023,unittestgen2025}.
Closer to our setting, skill-centric methods verify synthesized skills before use, either by synthesizing skills with verification at inference time \citep{skillgen2026} or by co-evolving skills together with a learned verifier \citep{coevoskills2026}.
These signals derive from the model's own priors or the target task itself, which limits calibration and risks supervision leakage.
\openskill's virtual tasks differ by anchoring verification to independently verifiable facts retrieved from the open world, yielding a practice environment that remains isolated from hidden target-task supervision.

\section{Conclusion}

We studied \emph{open-world self-evolution}, where an agent starts from only a task prompt and must build both its skills and its own verification signals from open-world resources, without target-task supervision during learning.
\openskill realizes this by acquiring grounded knowledge and verification anchors from the open world, synthesizing them into transferable skills, and refining those skills against self-built virtual tasks rather than target answers.
Across three benchmarks and two target agents it attains the best automated pass rate while honoring the no-supervision constraint, with skills that transfer to weaker models and a self-built verifier that aligns with ground-truth outcomes it never observes.
This points to open-world acquisition and leakage-free verification as a path toward agents that keep improving after deployment, where curated skills and reliable feedback are unavailable.

\section*{Limitations}

Open-world self-evolution introduces new challenges.
First, web and repository sources may be noisy, outdated, or contradictory.
The framework therefore requires provenance tracking and source validation.
Second, virtual tasks may fail to capture the full difficulty of real target tasks.
If virtual tasks are too easy, they may overestimate skill quality; if they are derived from hidden answers or verifier behavior, they may reintroduce target-task supervision.
Third, open-world research can increase cost and latency relative to closed-world skill generation.

\bibliographystyle{bibstyle}
\bibliography{custom}

\appendix
\section{Experimental Details}
\subsection{Model Details}
\label{app:models}

The \openskill pipeline uses models from three families. All LLM roles in
the pipeline are instantiated either with Anthropic Claude
(\texttt{claude-opus-4-6}),\footnote{\url{https://www.anthropic.com/news/claude-opus-4-6}}
or OpenAI
GPT (\texttt{gpt-5.2}).\footnote{\url{https://openai.com/index/introducing-gpt-5-2/}} The two open-world
retrieval roles use Google Gemini:\footnote{\url{https://ai.google.dev/gemini-api/docs/models/deep-research-pro-preview-12-2025}}
the main knowledge acquisition ($\mathcal{D}$) calls the Gemini Deep Research
agent (\texttt{deep-research-pro-preview-12-2025}), while verification-knowledge
retrieval ($\mathcal{D}^{v}$) and diagnostic-driven targeted retrieval use the
search-grounded \texttt{gemini-3.1-flash-lite}.\footnote{\url{https://ai.google.dev/gemini-api/docs/models/gemini-3.1-flash-lite-preview}} Zero-shot evaluation uses the \texttt{claude code}\footnote{\url{https://docs.claude.com/en/docs/claude-code/overview}} for Opus 4.6, \texttt{codex}\footnote{\url{https://github.com/openai/codex}} for GPT-5.2, and
\texttt{terminus-2} agent~\citep{merrill2026terminalbenchbenchmarkingagentshard} for other LLMs.
Version strings denote the exact model snapshots used in our experiments.

\subsection{Dataset Details}
\label{app:datasets}

We evaluate on three agentic benchmarks. \textsc{SkillsBench}
\citep{skillsbench2026} is our primary benchmark and spans 11 task domains:
Software, Office, Science, Media, Cybersecurity, Finance, Robotics, Energy,
Manufacturing, Health, and Math. It is constructed so that skill quality, rather
than base reasoning, is the limiting factor on task success.
\textsc{SocialMaze}~\citep{xu2025socialmaze} is a social-reasoning suite of six
subtasks---FTS, HRD, REFT, RDP, SGA, and UPI (broken out in
Appendix~\ref{app:socialmaze-detail}). \textsc{ScienceWorld}
\citep{wang2022scienceworld} is an interactive science-experiment environment.
On every benchmark the ground-truth test suite $\mathcal{T}^{\text{GT}}_i$ is
hidden during skill construction and consulted only at final evaluation.

\subsection{Baselines}
\label{app:baselines}

We compare \openskill against seven automated conditions and one reference
upper bound, all sharing the same target agents and hidden-test evaluation
protocol:
\begin{itemize}[leftmargin=*]
\item \textbf{No Skill} runs the target agent on the task with no skill artifact
provided. It isolates the agent's parametric competence and serves as the
zero-knowledge floor against which every skill-construction method is measured.

\item \textbf{Self-Gen} (Self-Generated Skills) reproduces the single-pass
self-generation condition of SkillsBench~\citep{skillsbench2026}. Drawing on its
parametric knowledge alone, the agent authors one to five \texttt{SKILL.md}
documents in a single forward pass and immediately uses them to solve the task.
There is no open-world retrieval, no interaction-trace mining, and no
evolution or verification loop, so the skills can only re-express knowledge the
model already holds. This baseline directly tests the SkillsBench observation
that models cannot reliably author the procedural knowledge they benefit from
consuming (prompt in Appendix~\ref{app:prompt-selfgen}).

\item \textbf{CoT} (CoT-Guided Self-Generation) preserves the single-pass,
parametric-only setting of Self-Gen but scaffolds the drafting with an explicit
five-step chain-of-thought prompt that walks the agent through analyzing the
task, recalling relevant procedures, structuring the skill, drafting it, and
self-reviewing before solving. It isolates whether structured reasoning over
existing knowledge---without any external information or feedback---is
sufficient to produce useful skills (prompt in Appendix~\ref{app:prompt-cot}).

\item \textbf{Skill Creator}~\citep{skill-creator} is Anthropic's official Claude
Code agent skill for authoring, evaluating, and iteratively improving skills. It
runs a \emph{Draft\,$\rightarrow$\,Test\,$\rightarrow$\,Review\,$\rightarrow$\,Improve\,$\rightarrow$\,Repeat}
loop: it captures intent, writes a \texttt{SKILL.md} with progressive
disclosure (metadata, body, and bundled scripts/references), executes
with-skill and baseline test cases, grades them with quantitative assertions,
and revises on the resulting feedback. Crucially, this refinement loop draws on
parametric knowledge and self-graded test cases rather than open-world sources
or a supervision-free verifier, so it iterates but does not acquire genuinely
new external knowledge.

\item \textbf{AutoSkill}~\citep{autoskill} is an experience-driven lifelong
learning framework that abstracts, maintains, and reuses skills from dialogue
and interaction traces, organizing them into a hierarchical skill bank
(domain\,$\rightarrow$\,family\,$\rightarrow$\,leaf skill) as a model-agnostic
plugin layer that injects relevant skills into future requests without
retraining. In our protocol we invoke the released AutoSkill toolkit to
construct a skill set per task from the task instruction and copy the generated
skills into the agent's skill directory. Because its self-evolution operates
over prior knowledge and accumulated traces rather than open-world retrieval,
it remains a closed-world refinement baseline.

\item \textbf{Memento}~\citep{zhou2026memento} is a memory-based reinforcement
learning framework in which reusable skills are stored as markdown files and
improved through a \emph{Read--Write Reflective Learning} mechanism: a skill
router selects relevant skills in the read phase, and the agent updates its
skill library from new experience in the write phase, with a background
consolidation (``dream'') step that compresses accumulated memory---all without
updating model parameters. We run the released system per task, excluding its
built-in utility skills and retaining only the newly created ones. Its updates
are driven by self-generated experience rather than external knowledge or a
supervision-free verifier.

\item \textbf{SkillNet}~\citep{liang2026skillnet} is an open infrastructure for
creating, evaluating, and connecting AI skills at scale, scoring skills along
five dimensions (safety, completeness, executability, maintainability, and
cost-awareness). Following its reported evaluation setting, we include SkillNet
on \textsc{ScienceWorld} only; the corresponding cells are left empty on the
other benchmarks.

\item \textbf{Human Curated} skills are the expert-authored reference skills
shipped with each benchmark. They serve as a reference upper bound and are
excluded from the best-automated-method comparison.
\end{itemize}

For the third-party methods that ship as executable systems---Skill Creator,
AutoSkill, and Memento---we use the released implementations with
\texttt{claude-opus-4-7} as the underlying backbone, the same target agents, and
the same hidden-test evaluation protocol as \openskill, so that differences in
reward reflect the skill-construction mechanism rather than the model or the
evaluation harness.

\subsection{Evaluation Metrics}
\label{app:eval-metrics}

For \textsc{SkillsBench}, we report the average reward (pass rate) over a benchmark's tasks, computed from
the hidden ground-truth test suite as in Eq.~\ref{eq:eval}. Each constructed
skill set is deployed under $n_{\text{eval}}=5$ independent zero-shot evaluation
runs (Appendix~\ref{app:selection}), and per-domain scores are averaged into the
overall figure reported in each table. For the virtual-verifier analysis
(Section~\ref{sec:surrogate_analysis}) we additionally report alignment
statistics---precision, recall, agreement, and association tests (Fisher's exact
test and point-biserial correlation)---and the coverage of ground-truth test
intents.

For \textsc{SocialMaze}, each of the six tasks is evaluated by
task-specific accuracy (\%): the fraction of test scenarios
in which the model produces a correct prediction
(e.g., identifying the spy, predicting the accept/reject
decision, exact-matching the star rating).
We report the macro-average across all six tasks.

For \textsc{ScienceWorld}, we use the environment completion
score (0--100) returned by the simulator, which measures
the degree to which the agent fulfills the task objective.
We report the mean score across all task variations.

For brevity, we refer to all metrics uniformly as
\emph{reward} throughout the paper.

\subsection{\openskill Hyperparameters}
\label{app:hparams}

Table~\ref{tab:hparams} lists the concrete configuration used in our
experiments.

\begin{table}[t]
\centering
\small
\begin{tabular}{ll}
\toprule
\textbf{Component} & \textbf{Setting} \\
\midrule
\multicolumn{2}{l}{\emph{Knowledge acquisition} ($\mathcal{D}$)} \\
\quad Deep-research agent & \texttt{deep-research-pro} \\
\quad Poll interval / hard timeout & 10\,s / 3600\,s \\
\quad Skills per plan & 1--4 \\
\midrule
\multicolumn{2}{l}{\emph{Verification knowledge} ($\mathcal{D}^{v}$)} \\
\quad Search model & \texttt{gemini-3.1-fl} \\
\quad Temperature & 0.3 \\
\quad ``Already known'' context & 4000 chars \\
\midrule
\multicolumn{2}{l}{\emph{Virtual testing \& evolution} ($g$, $\tilde{r}$)} \\
\quad Pass threshold to exit & $\tilde{r}=1.0$ \\
\quad Max refinement retries ($J$) & 3 \\
\quad Max host interventions & 1 \\
\quad Idle / stale episode limit & 3 / 10 \\
\quad Token-budget warn / stop & 70\% / 90\% \\
\quad Verifier / creator max episodes & 20 / 60 \\
\midrule
\multicolumn{2}{l}{\emph{Diagnostic-driven retrieval} ($\mathcal{D}$ on gaps)} \\
\quad Classifier model & \texttt{gemini-3.1-fl} \\
\quad Classifier temp. / max tokens & 0.1 / 200 \\
\quad Max searches per task & 3 \\
\midrule
\multicolumn{2}{l}{\emph{Zero-shot evaluation}} \\
\quad Evaluation runs $n_{\text{eval}}$ & 5 \\
\bottomrule
\end{tabular}
\caption{\openskill configuration used in experiments. Model names are
abbreviated; see Appendix~\ref{app:impl} for the exact identifiers and the
role of each component.}
\label{tab:hparams}
\end{table}

\section{Pipeline Implementation Details}
\label{app:impl}

This appendix specifies how the abstract functions in
Section~\ref{sec:method}---the open-world retrieval $\mathcal{D}$ and
$\mathcal{D}^{v}$, the virtual-test generator $g$, the refinement loop, and
the gap/bug diagnosis---are realized, so that the pipeline is reproducible.
The pipeline is orchestrated on the host, while skill creation and
verification run inside a per-task Docker container.

\subsection{Open-World Retrieval $\mathcal{D}$}
\label{app:retrieval}

$\mathcal{D}$ is decomposed into three host-side stages.

\paragraph{Query synthesis.} An LLM \emph{Agent Planner} reads the task
instruction $\mathcal{I}_i$, the task metadata (category, tags, difficulty),
and a preview of the environment files in $\mathcal{E}_i$ (text files truncated
to 2000 characters; PDFs extracted to 5000 characters; \texttt{skills/} and
\texttt{doc/} directories excluded). It emits a single free-text research query
that requests library APIs, function signatures, parameter defaults, working
code examples, and common pitfalls, while being explicitly instructed not to
include the solution approach.

\paragraph{Leakage filtering.} Before any query is issued, a
\texttt{sanitize\_query} step removes the benchmark name and its spelling
variants (case-insensitive) from the query string, so that the retrieval
engine cannot match the benchmark's own pages and leak target answers. This
filter is applied to every query issued by $\mathcal{D}$ and $\mathcal{D}^{v}$.

\paragraph{Retrieval and planning.} The sanitized query is submitted to a
commercial deep-research agent (\texttt{deep-research-pro-preview-12-2025})
through an asynchronous interactions API: a single query is submitted and
polled every 10\,s up to a 3600\,s hard timeout; the multi-step web search and
synthesis are performed server-side. The result is written as a background
document $k_i$ together with a deduplicated list of source URLs. A second LLM,
the \emph{Skill Planner}, then decomposes $k_i$ into 1--4 skills (each with a
name, responsibility, key functions, and the background sections it depends
on) to form the plan $p_i$, and slices the background document into
per-skill reference material by fuzzy-matching section headers (falling back
to the full document when no section matches).

\subsection{Verification-Knowledge Retrieval $\mathcal{D}^{v}$}
\label{app:surrogate}

$\mathcal{D}^{v}$ is a second, orthogonal retrieval pass implemented as a
single search-grounded generation call (\texttt{gemini-3.1-flash-lite},
temperature 0.3, Google-Search grounding). Unlike $\mathcal{D}$, its prompt
does not ask for API knowledge; it asks only for four classes of
independently checkable anchors: (i) reference values that can be computed by
hand for small, well-known inputs; (ii) dataset-level statistical invariants
(row counts, sum-to-one constraints, monotonicity, value ranges); (iii)
cross-validation procedures using alternative tools or libraries; and (iv)
published benchmarks or reference implementations with known input--output
pairs. To keep $k_i^{v}$ disjoint from $k_i$, the first 4000 characters of the
background document are inserted into the prompt under an ``Already Known---do
not repeat'' block. The result $k_i^{v}$ is stored and later injected into the
virtual-test generator.

\subsection{Virtual-Test Generation $g$}
\label{app:verifier}

The virtual test suite $\tilde{\mathcal{T}}_i$ is produced by an
\emph{Independent Verifier}: a fresh LLM session that shares the container
(and therefore the produced output files) with the skill creator but
\emph{not} its conversation, reasoning, or implementation code. This
isolation is by design, to avoid confirmation bias---the verifier judges
outputs without seeing how they were generated.

The verifier is instructed to derive expected values either by independently
recomputing them from the environment inputs and the documented task rules, or
directly from the verification knowledge $k_i^{v}$, which is injected into its
prompt as the ``primary oracle'' of externally verified values. It emits a
\texttt{pytest} script of deterministic equality assertions (e.g.,
\texttt{assert x == y}) that is executed in the container. Each assertion
$\tilde{t}_{i,k}\in\{0,1\}$ is thus an exact, reproducible check. We emphasize
that the hidden test suite $\mathcal{T}^{\text{GT}}_i$ is never referenced: its
file paths are not provided to the verifier, and the SkillWeaver loop never
invokes the ground-truth oracle during construction. The isolation is enforced
at the process and prompt level (a separate session with no GT paths) rather
than by a filesystem sandbox.

A non-LLM parser reads the \texttt{pytest} output and computes the virtual
pass rate $\tilde{r}^{(j)}$ as $\text{passed}/\text{total}$ (skipped tests
excluded), yielding the quality signal in Eq.~\ref{eq:virtual-pass-rate}.
Across refinement rounds the verifier inherits its previous script and failure
list, repairing broken test code before adding deeper assertions, and is
capped at 60 tests to favor correctness over quantity.

\subsection{Iterative Refinement and Termination}
\label{app:loop}

Refinement is performed by the skill creator itself, in a single long-lived
session: when $\tilde{r}^{(j)}<1$, the host returns a structured failure
report---passed/total counts, the failed-assertion list, and an optional
diagnosis---and instructs the creator to fix the logic in its skill scripts
and regenerate outputs (never to edit outputs directly). The loop exits with
status \texttt{surrogate\_pass} only when $\tilde{r}=1.0$ and auxiliary
structural checks pass. The bound $J$ in Section~\ref{sec:stage2} is realized
as a cap of 3 surrogate-failure retries (with at most one successful
intervention); additional early-stopping guards trigger on idle/stale
episodes (3/10) and on token-budget exhaustion (warning at 70\%, stop at
90\%).

\subsection{Gap-vs-Bug Diagnosis and Targeted Retrieval}
\label{app:diagnosis}

When a refinement round fails, an LLM classifier
(\texttt{gemini-3.1-flash-lite}, temperature 0.1, 200 max tokens) reads the
failed assertions, the diagnosis, and the prior domain knowledge, and emits
one line: \texttt{SELF-FIXABLE} (an implementation bug---wrong variable,
off-by-one, format mismatch, type error, or a fix already implied by existing
knowledge) or \texttt{NEEDS-DR} (a knowledge gap---an unknown correct value,
an algorithm/parameter choice requiring domain expertise, or library usage not
covered by prior knowledge). On \texttt{SELF-FIXABLE}, only the verifier
feedback is returned and the creator fixes the code unaided. On
\texttt{NEEDS-DR}, a targeted search $k_i^{(\text{gap})}$ is issued (sharing
the search-grounding path of $\mathcal{D}^{v}$) and its result is appended to
the feedback. Targeted retrieval is capped at 3 searches per task, with
already-searched topics listed in the query to discourage repetition; if the
classifier call fails, the system conservatively defaults to
\texttt{SELF-FIXABLE} (no retrieval).

\subsection{Final Skill Selection}
\label{app:selection}

Because the creator edits one skill set in place across rounds, the final
skill set $\hat{\mathcal{S}}^*_i$ deployed in Stage~3 is the state of the
\texttt{evo-*} skills at loop termination---the most recently refined
version---rather than a best-of-$N$ snapshot selected by virtual pass rate.
These skills are exported from the container and copied into the target
workspace, where the agent uses them under $n_{\text{eval}}=5$
independent zero-shot evaluation runs scored by $\mathcal{T}^{\text{GT}}_i$.

\section{Failure Modes of Virtual Verifier}
\label{app:failure-modes}
We analyze the 33 disagreement cases between the virtual
verifier and GT evaluation (25 false positives and 8 false
negatives) to characterize the failure modes of the
virtual verifier.

\paragraph{False positives (proxy pass, GT fail).}
The 25 false positive tasks fall into three categories.
(1)~\emph{High-accuracy near-misses} (12 tasks, mean
acc\,=\,0.81): these tasks pass the majority of GT tests
but receive zero reward due to the strict all-or-nothing
evaluation criterion (reward\,>\,0 requires every test to
pass across all runs). The virtual verifier correctly
identifies that the skill produces largely correct outputs,
but cannot anticipate the few remaining edge-case failures
that the GT suite catches.
(2)~\emph{Partial correctness} (11 tasks, mean acc\,=\,0.52):
the surrogate tests are satisfied by outputs that are
structurally valid but semantically incomplete---for instance,
producing a well-formed JSON with incorrect numerical values.
This reflects the virtual verifier's reliance on format-level
and boundary checks, which are insufficient for tasks requiring
deep computational verification.
(3)~\emph{Genuine misalignment} (2 tasks, acc\,=\,0.0):
the surrogate tests fail to capture any aspect of the correct
behavior, approving entirely incorrect outputs.
Both cases involve domain-specific chemical similarity search
and code translation tasks where the virtual verifier lacks
the prerequisite knowledge to generate meaningful test oracles.

\paragraph{False negatives (proxy fail, GT pass).}
The 8 false negative tasks cluster into two patterns.
(1)~\emph{Near-pass} (3 tasks): the virtual verifier achieved
89--97\% surrogate pass rate but fell short of 100\%, typically
failing on one or two overly strict surrogate assertions.
These tasks succeed on GT evaluation because the unmet
surrogate tests target edge cases that are not tested by the
GT suite.
(2)~\emph{Verifier infrastructure failure} (5 tasks):
the surrogate test generation or execution pipeline failed
entirely (0\% pass rate), yet the agent still produced a
correct solution. These tasks are predominantly non-standard
(CVE patches, build system fixes, proof assistants)
where the virtual verifier's pytest-based testing framework
is fundamentally unsuitable for validating the output.

\section{SocialMaze Per-Subtask Results}
\label{app:socialmaze-detail}

Table~\ref{tab:socialmaze-detail} expands the SocialMaze averages of
Table~\ref{tab:additional-benchmarks} into the six SocialMaze subtasks (FTS,
HRD, REFT, RDP, SGA, and UPI) for both target agents. Per-subtask scores vary
widely across methods; \openskill attains the best overall average on
both target agents (82.7\% on Opus, 70.7\% on GPT), driven largely by the
harder reasoning subtasks (REFT, UPI on Opus; REFT, RDP on GPT) rather than
uniform gains across all subtasks.

\begin{table}[t]
\centering
\scriptsize
\setlength{\extrarowheight}{1.5pt}
\setlength{\tabcolsep}{3pt}
\begin{tabular}{l rrrrrr | r}
\toprule
Method & FTS & HRD & REFT & RDP & SGA & UPI & Avg \\
\midrule
\rowcolor{tendergreen}
\multicolumn{8}{l}{\textbf{Opus 4.6}} \\
\rowcolor{tenderblue}
\openskill & 98.0 & 90.0 & 53.7 & 80.0 & 100.0 & 74.5 & \textbf{82.7} \\
No Skill      & 96.0  & 91.2 & 50.7 & 80.0 & 100.0 & 71.5 & 81.6 \\
Skill Creator & 98.0  & 91.2 & 49.6 & 77.5 & 100.0 & 70.0 & 81.0 \\
Self-Gen      & 92.0  & 92.5 & 51.1 & 75.0 & 100.0 & 72.0 & 80.4 \\
Memento       & 100.0 & 91.2 & 46.7 & 75.0 & 98.8  & 70.0 & 80.3 \\
CoT           & 94.0  & 88.8 & 52.3 & 70.0 & 100.0 & 71.5 & 79.4 \\
AutoSkill     & 100.0 & 92.5 & 27.3 & 75.0 & 100.0 & 68.5 & 77.2 \\
\midrule
\rowcolor{tendergreen}
\multicolumn{8}{l}{\textbf{GPT 5.2}} \\
\rowcolor{tenderblue}
\openskill & 98.0 & 60.0 & 54.6 & 62.5 & 80.0 & 69.0 & \textbf{70.7} \\
Skill Creator & 100.0 & 72.5 & 39.3 & 55.0 & 86.2 & 65.5 & 69.8 \\
Self-Gen      & 98.0  & 67.5 & 43.7 & 55.0 & 68.8 & 72.5 & 67.6 \\
Memento       & 100.0 & 71.2 & 46.3 & 57.5 & 63.7 & 65.5 & 67.4 \\
CoT           & 94.0  & 58.8 & 51.9 & 55.0 & 70.0 & 70.5 & 66.7 \\
No Skill      & 96.0  & 52.5 & 51.0 & 57.5 & 72.5 & 70.0 & 66.6 \\
AutoSkill     & 98.0  & 67.5 & 29.7 & 55.0 & 72.5 & 72.5 & 65.9 \\
\bottomrule
\end{tabular}
\caption{Per-subtask reward (\%) on SocialMaze for the Opus~4.6 and GPT~5.2
target agents. \openskill rows are shaded; best average per target
agent in \textbf{bold}.}
\label{tab:socialmaze-detail}
\end{table}

\section{Computational Cost}
\label{sec:cost}

Table~\ref{tab:cost} reports the per-task computational cost of \textsc{OpenSkill} on \textsc{SkillsBench} (84 tasks, Opus 4.6). The pipeline consists of two phases: \emph{skill creation} (Stages 1--4: deep research, skill planning, and skill synthesis with a virtual verifier loop) and \emph{evaluation} (Stage 5: 5 independent agent runs using the generated skill).

\begin{table}[t]
\centering
\scriptsize
\setlength{\tabcolsep}{4pt}
\resizebox{0.4\columnwidth}{!}{%
\begin{tabular}{llrr}
\toprule
\textbf{Stage} & \textbf{Description} & \textbf{Tokens} & \textbf{Time} \\
\midrule
1 & Deep Research (Gemini) & $\sim$10K & $\sim$8 min \\
2 & Skill Plan Generation  & $\sim$10K & $<$1 min \\
3 & Surrogate DR (Gemini)  & $\sim$2K  & $\sim$2 min \\
4 & Skill Creation + VV Loop & 727K & 29.0 min \\
\cmidrule{2-4}
  & \textbf{Skill creation total} & \textbf{$\sim$749K} & \textbf{$\sim$39 min} \\
\midrule
5 & GT Evaluation (5 runs) & $\sim$400K & 91.5 min \\
\midrule
  & \textbf{End-to-end total} & \textbf{$\sim$1.14M} & \textbf{$\sim$131 min} \\
\bottomrule
\end{tabular}%
}
\caption{Per-task computational cost on \textsc{SkillsBench} (Opus 4.6). Token counts for Stage 4 are measured from execution logs; others are estimated from artifact sizes. Stages 1 and 3 use Gemini 2.5 Flash; all other stages use Opus 4.6.}
\label{tab:cost}
\end{table}

Skill creation dominates the token budget (749K out of 1.14M tokens, 66\%) but accounts for only 30\% of the wall-clock time (39 out of 131 minutes). This discrepancy arises because the virtual verifier loop (Stage 4, averaging 3.0 iterations) runs as a single sequential LLM session, whereas ground-truth (GT) evaluation requires 5 independent, Docker-containerized agent runs. Across all 84 tasks, the total skill creation cost accumulates to 31.4 API-hours and $\sim$47M tokens. The full end-to-end pipeline (creation and evaluation) totals 140 hours and $\sim$97M tokens, corresponding to an estimated API cost of $\sim$\$1,800 at Opus 4.6 list pricing (\$15/M input, \$75/M output).

Importantly, skill creation is a \emph{one-time} cost: once generated, skills are reused across models and runs without additional creation overhead. In the cross-model transfer experiments, skill-equipped evaluation takes 16--27 minutes per task depending on the target model (Haiku: 17.7 min, DeepSeek: 18.1 min, Qwen 3 Coder: 15.7 min, Mistral: 27.0 min), incurring zero additional skill creation cost.

\paragraph{Computational Cost Comparison.}
Table~\ref{tab:cost_analysis} compares the per-task, per-run evaluation cost between the No-Skill baseline and \textsc{OpenSkill}. Both configurations use Opus 4.6 as the backbone LLM under the same evaluation harness.

The No-Skill agent spends an average of 465.0~s per evaluation run. Considering the evaluation stage alone, \textsc{OpenSkill} requires a median of 368.2~s per run, which is \emph{comparable} with the No-Skill baseline. The substantial gap between the eval-only mean (845.4~s) and median (368.2~s) is driven by a small number of long-running tasks.

\begin{table}[t]
\centering
\scriptsize
\begin{tabular}{lcc}
\toprule
Result & \textbf{No-Skill} & \textbf{OpenSkill} \\
\midrule
Per-run time, mean (s) & 465.0  & 845.4 \\
Per-run time, median (s) & 347.6 & 368.2 \\
Mean reward & 25.5\% & \textbf{43.6\%} \\
\bottomrule
\end{tabular}
\caption{Per-run evaluation cost on \textsc{SkillsBench} (Opus 4.6 backbone). It reports GT evaluation time, excluding skill creation.}
\label{tab:cost_analysis}
\end{table}

\section{Information Isolation Audit}
\label{app:leakage_audit}

A key design requirement of the Virtual Verifier is strict information isolation:
it must generate surrogate tests \emph{without} access to ground-truth (GT) tests,
solutions, or the skill creator's internal state.
We verify this through a four-layer audit.

\paragraph{Code-level isolation.}
The base agent accepts exactly two inputs:
the task instruction and environment files (input data, Dockerfile).
Its function signature explicitly excludes solution and test paths.
The independent verifier (\texttt{independent\_verifier.py})
instantiates a \emph{separate} LLM session with no shared context from the skill creator,
preventing any cross-agent information leakage.

\paragraph{Container-level isolation.}
Each task runs inside a Docker container built from the task's \texttt{environment/Dockerfile}.
The GT test directory (\texttt{tests/}) and solution directory (\texttt{solution/})
reside on the host filesystem as siblings to \texttt{environment/}
and are \emph{never} mounted or copied into the container.
The verifier agent can only observe files under \texttt{/app/environment/}
and outputs written by the creator agent.

\paragraph{GT oracle bypass.}
The OpenSkill evolution loop overrides the parent class's GT oracle,
ensuring the GT oracle is never invoked during skill creation.
The loop exits upon surrogate test passage,
without ever reaching the GT evaluation code path.
The only GT-derived signal is a single pass/fail bit
used when the verifier is re-invoked to write deeper tests;
no GT test content, assertion details, or expected values are exposed.

\paragraph{Log-level verification.}
We audited \texttt{evolution\_run\_log.json} files.
Zero references to GT test files
appear in agent execution trajectories.
The only GT-related entries reside in post-hoc pipeline evaluation fields
that are recorded \emph{after} the agent has exited and are never fed back to any agent.

\paragraph{Prompt-level enforcement.}
Both the surrogate writer and independent verifier system prompts
contain explicit instructions:
\emph{``You must ONLY use information from the task instruction and environment files.
You have NO access to the solution or ground-truth tests.''}

\section{Baseline Prompts}
\label{app:prompts}

\subsection{Self-Generated Skills Prompt}
\label{app:prompt-selfgen}

This prompt replicates the self-generation condition of
SkillsBench~\citep{skillsbench2026} (Appendix~C.6). It is appended to the task
instruction; the agent generates skills in-session before solving the task,
with no external verification.

\begin{promptbox}{Self-Generated Skills Prompt}
\ttfamily\small
Important: Generate Skills First\\
Before attempting to solve this task:
\begin{enumerate}[leftmargin=2.4em, itemsep=2pt, topsep=3pt, parsep=0pt]
\item Analyze the task requirements and identify what domain knowledge, APIs, or techniques are needed.
\item Write 1--5 modular skill documents.
\item Save each skill as a markdown file in environment/skills/.
\item Then solve the task using the skills you created as reference.
\end{enumerate}
\end{promptbox}

\subsection{CoT-Guided Self-Generation Prompt}
\label{app:prompt-cot}

This prompt extends the Self-Generated Skills baseline with a structured
five-step chain-of-thought workflow. Despite the added structure, the agent
still lacks external verification feedback, and this condition achieves only
30.7\% pass rate (comparable to the no-skill baseline).

\begin{promptbox}{CoT-Guided Self-Generation Prompt}
\ttfamily\small
Step 1: Task Analysis -- identify domain, tools, output format, pitfalls\\
Step 2: Skill Architecture Design -- plan 1--3 focused skills\\
Step 3: Write Skills with Progressive Disclosure
\begin{enumerate}[label=(\alph*), leftmargin=2.6em, itemsep=2pt, topsep=3pt, parsep=0pt]
\item YAML frontmatter: name and description
\item Key constraints and rules
\item Step-by-step workflow with decision points
\item Common mistakes to avoid and edge cases
\item If helpful, include scripts/ with reusable utility code
\end{enumerate}
Step 4: Self-Verify -- re-read instruction, check every requirement has coverage\\
Step 5: Execute
\end{promptbox}

\end{document}